\documentclass[runningheads]{llncs}

\usepackage{subfigure}
\usepackage{bbding}
\usepackage[misc]{ifsym}
\usepackage{graphicx}
\usepackage{booktabs} 
\usepackage{amsmath,amssymb,amsfonts}
\usepackage{algorithm}  
\usepackage{algorithmicx}  
\usepackage{algpseudocode}  
\floatname{algorithm}{Algorithm}  

\begin{document}
\title{Multi-Objective Pruning for CNNs Using Genetic Algorithm}
\titlerunning{Multi-Objective Pruning for CNNs Using Genetic Algorithm}
\author{Chuanguang Yang\inst{1,2} \and
 Zhulin An\inst{1(}\Envelope\inst{)} \and Chao Li\inst{1} \and Boyu Diao\inst{1} \and
Yongjun Xu\inst{1}}
\authorrunning{C. Yang et al.}
%
\institute{Institute of Computing Technology,  Chinese Academy of Sciences, \\ Beijing 100190, China \and
University of Chinese Academy of Sciences, Beijing 100049, China
\email{\{yangchuanguang,anzhulin,lichao,diaoboyu2012,xyj\}@ict.ac.cn}}
\maketitle              
\begin{abstract}
In this work, we propose a heuristic genetic algorithm (GA) for pruning convolutional neural networks (CNNs) according to the multi-objective trade-off among error, computation and sparsity. In our experiments, we apply our approach to prune
pre-trained LeNet across the MNIST dataset, which reduces 95.42\% parameter size and achieves 16× speedups of convolutional layer computation with
tiny accuracy loss by laying emphasis on sparsity and computation, respectively. Our empirical study suggests that GA is an alternative pruning approach for obtaining a competitive compression performance. Additionally, compared with state-of-the-art approaches, GA can automatically pruning CNNs based on the multi-objective importance by a pre-defined fitness function.

\keywords{Genetic algorithm  \and Convolutional neural networks \and Multi-objective pruning.}
 
\end{abstract}
\section{Introduction}
 Vision application scenarios often have different requirements in terms of multi-objective importance about error, computational cost and storage for convolutional neural networks (CNNs), but state-of-the-art pruning approaches
  do not take this into account. Thus, we develop the genetic algorithm (GA) that can iteratively prune redundant parameters based on the multi-objective trade-off by a two-step procedure. First, we prune the network by taking the advantages of swarm intelligence. Next, we retrain the elite network and reinitialize the population by the trained elite. Compared with state-of-the-art approaches, our approach obtains a comparable result on sparsity and a significant improvement on computation reduction. In addition, we detail how to adjust the fitness function for obtaining diverse compression performances in practical applications.

\section{Proposed Approach}

\subsection{Evaluation Regulation}
 Similar to general evolutionary algorithms, we design a fitness function $f$ to evaluate the comprehensive performance of a genome. In our method, $f$ is defined by the weighted average of error rate $e$, computation remained rate $c$ and sparsity $s$. And our target is to minimize the fitness function as follows:
\begin{equation}
\begin{aligned}
&\min f =  \min(\lambda_{1}e+\lambda_{2}c+\lambda_{3}(1-s)) \\
&s.t.\ 0\leqslant e,c,s,\lambda_{1},\lambda_{2},\lambda_{3} \leqslant 1,\lambda_{1}+\lambda_{2}+\lambda_{3}=1\\
\end{aligned}
\end{equation}

The coefficients $\lambda_{1}$,\ $\lambda_{2}$, and $\lambda_{3}$ adjust the importance of the three objectives. $e$, $c$ and $s$ denote the percentage of misclassified samples, remained  multiplication-addition operations (FLOPs) and zeroed out parameters, respectively. From the experimental analyses in section \ref{exp}, treating the multi-objective nature of the problem by linear combination and scalarization is indeed effective and consistent to our expectation, albeit more sophisticated fitness function may further improve the results.
\begin{figure}[tbp] 
	\centering 
	\subfigure[Filter-wise pruning.]{
		\label{Fig.sub.1} 
		\includegraphics[width=0.63\textwidth]{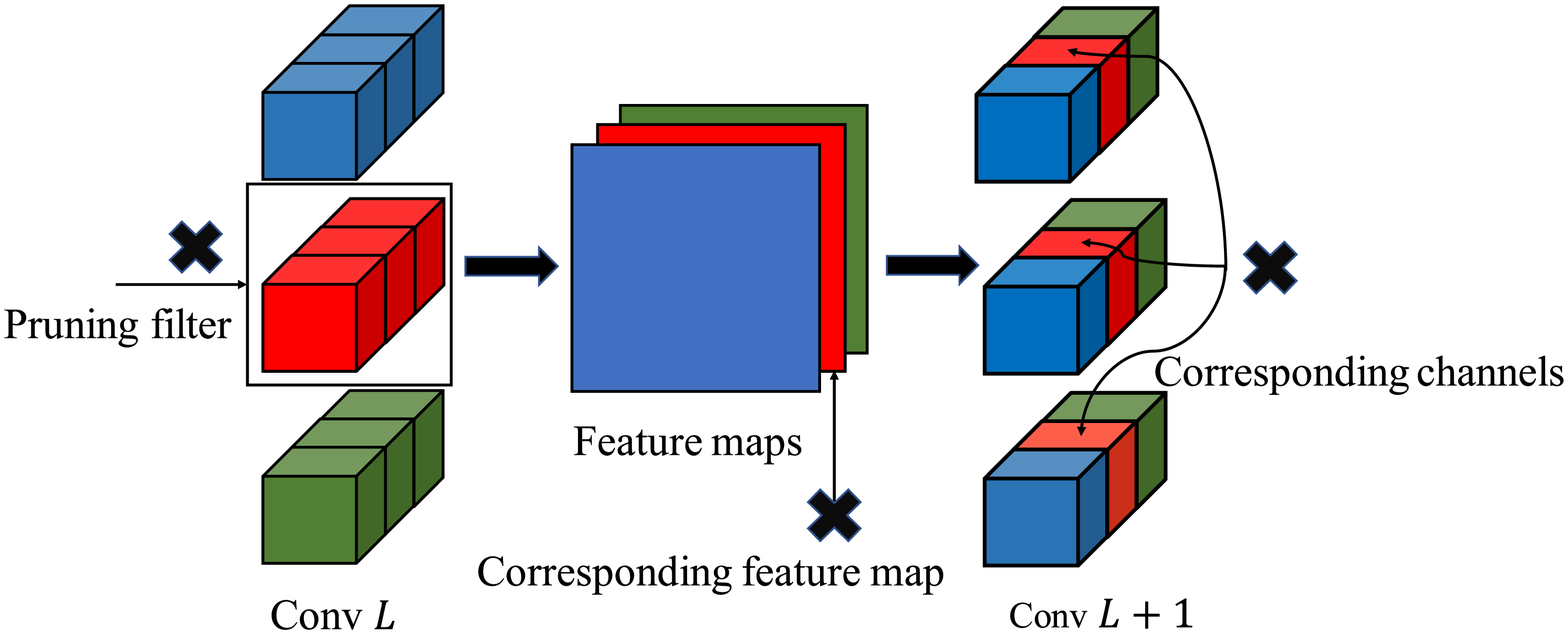}} 
	\subfigure[Connection-wise pruning.]{ 
		\label{Fig.sub.2} 
		\includegraphics[width=0.33\textwidth]{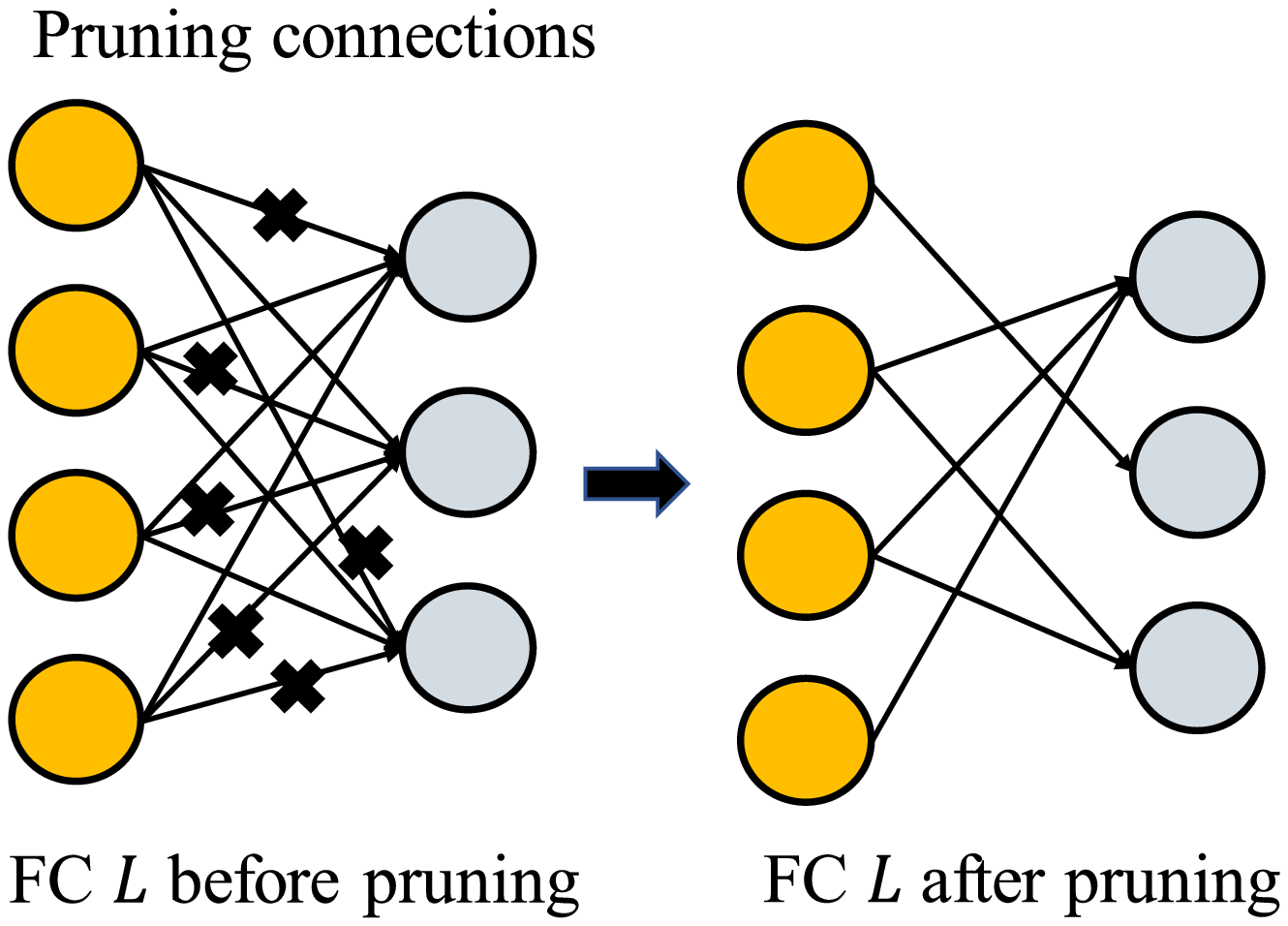}} 
	\caption{Pruning techniques of CONV layer (a) and FC layer (b) in mutation phase.  For the current CONV layer, we carry out a filter-wise pruning based on mutation rate $P_{mc}$, and then a corresponding channel-wise pruning will also take place for the next CONV layer. For the current FC layer, we carry out a connection-wise pruning based on mutation rate $P_{mf}$. } 
	\label{Fig.lable} 
\end{figure} 
\subsection{Heuristic Pruning Procedure}
\subsubsection{Genetic Encoding and Initialization.} A CNN is encoded to a genome including $M$ parameter genes that denoted by $\theta^{1}, \theta^{2},...,\theta^{M}$, where $M$ denotes the depth of the CNN, $\theta^{m}$ denotes the $m$th layer parameter with a 4D tensor of size $F_{m}$×$C_{m}$×$H_{m}$×$W_{m}$ in convolutional (CONV) layer, or a 2D tensor of size $O_{m}$×$I_{m}$ in fully-connected (FC) layer, where $F_{m}$, $C_{m}$, $H_{m}$ and $W_{m}$ denote the size of filters, input channels, height and width of kernels, $O_{m}$ and $I_{m}$ denote the
size of output and input features, respectively. We apply $N$ times mutations on a pre-trained CNN to generate the initial population consisting of $N$ genomes. 
 
\subsubsection{Selection.}
We straightforward select the top $K$ genomes with minimum fitness to reproduce next generation. It is worth mentioning that we have attempted a variety of selection operations, such as tournament selection, roulette-wheel selection and truncation selection. Our empirical results indicate that different selection operations finally obtain the similar performance but the vanilla selection which we adopt has the fastest convergence speed.

\subsubsection{Crossover.}
Crossover operations are occurred among the selected genomes based on the crossover rate $P_{c}$. We employ the classical microbial  crossover which is first proposed in \cite{microbial} inspired by bacterial conjugation. For each crossover, we choose two genomes randomly, from which the one with lower fitness is called Winner genome, and the other one is called Loser genome. Then, each gene in Loser genome is copied from Winner genome based on 50\% probability. Thus, Winner genome can remain unchanged to preserve the good performance, and  Loser genome can be modified to generate possibly better performance by the infection of Winner genome. One potential strength of microbial  crossover is implicitly remaining the elite genome to the next generation, since the fittest genome can win any tournaments against any genomes.

\subsubsection{Mutation.} Mutation performs for every genome except for the elite with mutation rate $P_{mc}$ and $P_{mf}$ in each CONV layer and FC layer, respectively. Follow \cite{Exploring}, we employ the coarse-grained pruning on CONV layers and fine-grained pruning on FC layers, both of which are sketched in Fig.\ref{Fig.lable}.

\begin{algorithm} 
	\label{alg}
	\caption{Multi-objective Pruning by GA}  
	\begin{algorithmic}[1] 
		\Require pre-trained CNN parameter $\theta_{initial}$, maximum number of iterations $G$, population size $N$, number
		of selected genomes $K$, crossover rate $P_{c}$, mutation rate $P_{mc}$
		and $P_{mf}$, number of interval iterations $T$
		\Ensure parameter of elite genome $\hat{\theta}^{G}$
		
		\For{$i = 1 \to N$}
		\State $\mathcal{P}_{i}^{g=0} \gets  \Call{mutation}{\theta_{initial},P_{mc}, P_{mf}}$
		\EndFor
		\For{$g = 1 \to G$} 
		\State $\mathcal{F}_{1,...,N}^{g-1} \gets f(\mathcal{P}^{g-1}_{1,...,N})$
		\State $elite \gets argmax_{i \in \{1,...,N\}} \mathcal{F}^{g-1}_{i}$
		\State $\hat{\theta}^{g-1} \gets \mathcal{P}^{g-1}_{elite}$
		\State $\mathcal{P}^{g}_{1,...,K} \gets \Call{selection}{\mathcal{P}^{g-1}_{1,...,N},\mathcal{F}_{1,...,N}^{g-1},K}$
		\State $\mathcal{P}^{g}_{1,...,N-1} \gets \Call{crossover}{\mathcal{P}^{g}_{1,...,K},P_{c}}$
		\State $\mathcal{P}^{g}_{1,...,N-1} \gets \Call{mutation}{\mathcal{P}^{g}_{1,...,N-1},P_{mc}, P_{mf}}$
		\State $\mathcal{P}^{g}_{1,...,N}\gets \mathcal{P}^{g}_{1,...,N-1} \cup \{\hat{\theta}^{g-1}\}$
		
		\If {$mod(g,T) = 0$}
		\State $\hat{\theta}^{g}\gets argmax_{\theta \in \mathcal{P}^{g}_{1,...,N}}f(\theta)$
		\State $\hat{\theta}^{g} \gets train(\hat{\theta}^{g})$
		\For{$i = 1 \to N-1$}
		\State $\mathcal{P}_{i}^{g} \gets  \Call{mutation}{\hat{\theta}^{g},P_{mc}, P_{mf}}$
		\EndFor
		\State $\mathcal{P}^{g}_{1,...,N}\gets \mathcal{P}^{g}_{1,...,N-1} \cup \{\hat{\theta}^{g}\}$
		\EndIf 
		\EndFor
		\State $\hat{\theta}^{G} \gets train(\hat{\theta}^{G})$

	\end{algorithmic}  
\end{algorithm}

\subsubsection{Main Procedure.} After each heuristic pruning process including selection, crossover and mutation with $T$ iterations, we retrain the elite genome so that the remained weights can compensate for the loss of accuracy, and then reinitialize the population by the trained elite genome. The above procedures are repeated iteratively until the fitness of the elite genome is convergence. Algorithm 1 illustrates the whole procedures of multi-objective pruning by GA.

%

\section{Experimental Results and Analyses}
\label{exp}
\begin{figure}[htbp]  
	\centering  
	\includegraphics[width=1.0\linewidth]{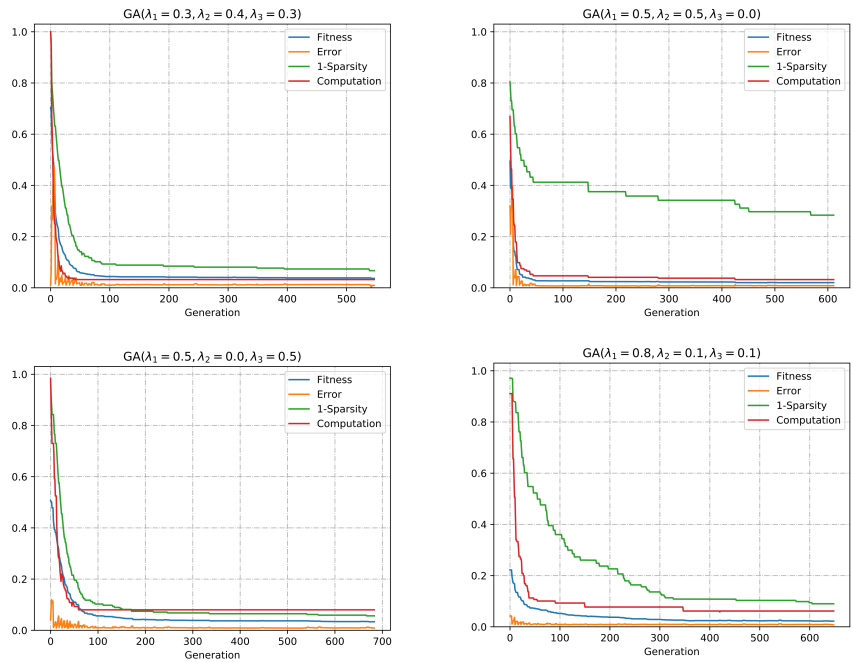}
	\caption{Pruning process of GA with different $\lambda_{1}\sim \lambda_{3}$. The blue, orange, green, red curves reflect the indicator of fitness, error, sparsity and FLOPs of the elite, respectively}  
	\label{fig:figure1}
\end{figure}
The hyper-parameter settings of GA are as follows: population size $N=30$, number
of selected genomes $K=5$, crossover rate $P_{c}=0.6$, mutation rate $P_{mc}=0.1$
and $P_{mf}=0.15$, iteration number $T=5$. Albeit we find that further hyper-parameter tuning can obtain better results, such as increasing population size or diminishing mutation rate, but corresponding with more time cost. 

Comprehensive comparison with state-of-the-art approaches
 is summarized in Table \ref{tab1}. We highlight in particular that different pruning performances can be obtained by adjusting $\lambda_{1}\sim \lambda_{3}$. Meanwhile, we empirically analyze the effectiveness by custom $\lambda_{1}\sim \lambda_{3}$ with corresponding curves which are exhibited in Fig.\ref{fig:figure1}. Note that CONV layers and FC layers are the main source of computation and parameter size, respectively. And $\lambda_{1}$ cannot be set too tiny in order to ensure the low error.

\begin{enumerate}
	\item $\mathbf {\lambda_{1}=0.3,\lambda_{2}=0.4,\lambda_{3}=0.3}$. With the approximate weights for $\lambda_{1}\sim \lambda_{3}$ as our baseline, which reach the overall optimal compression performance but with relatively higher error rate. 
	\item $\mathbf {\lambda_{1}=0.5,\lambda_{2}=0.5,\lambda_{3}=0}.$ This setting aims at high-speed inference for CNN. In this case, computation achieves maximum reduction, but sparsity is hard to optimize because GA pays less attention to pruning FC layers which are not the main source of computation.
	\item $\mathbf {\lambda_{1}=0.5,\lambda_{2}=0,\lambda_{3}=0.5}.$ This setting aims at a CNN with low storage. In this case, we obtain the utmost sparsity and high-level computation reduction simultaneously. Albeit CONV layers only play an unimportant role in the overall parameter size, it can also obtain the high-level sparsity because of the tractability with coarse granularity pruning. Thus, $\lambda_{3}$ can also indirectly facilitate computation reduction.
	\item $\mathbf {\lambda_{1}=0.8,\lambda_{2}=0.1,\lambda_{3}=0.1}.$ This setting aims at minimal performance loss. In this case, error curve is always at the low level resulting in that GA is conservative to pruning both CONV and FC layers. Hence, parameter and FLOPs curves are slower to fall compared with baseline. 
\end{enumerate}

\begin{table}
\caption{Comparison against the pruning approaches
	 evaluated on MNIST dataset\cite{data}. Note that bold entries represent the emphases on objectives laid by GA.}\label{tab1}
\centering
\renewcommand\arraystretch{1.2}
\begin{tabular}{|l|l|l|l|l|}
\hline
Approach
&Error:$e$ &Computation:$c$ &Sparsity:$s$& Accuracy change\\
\hline
	LeNet Baseline \ \cite{ref_url2,LeNet}
&\ 0.8\%\ &\ 100\%&\ 0\% &\ -\\
\hline
LNA\ \cite{LNA} &\ 0.7\%\ &\ -&\ 90.5\% &\ +0.1\%\\
SSL\ \cite{SSL} &\ 0.9\%\ &\ 25.64\%&\ 75.1\% &\ -0.1\%\\
TSNN\ \cite{TSNN} &\ 0.79\%\ &\ 13\%&\ 95.84\%&\ +0.01\%\\
SparseVD\ \cite{SparseVD} &\ 0.75\%\ &\ 45.66\%&\ 92.58\%&\ +0.05\%\\
StructuredBP\ \cite{StructuredBP} &\ 0.86\%\ &\ 9.53\%&\ 79.8\%&\ -0.06\%\\
$l_{0}$ Regularization\ \cite{l0s} &\ 1.0\%\ &\ 23.22\%&\ 99.14\%&\ -0.2\%\\
RA-2-0.1\ \cite{RA} &\ 0.9\%\ &\ -&\ 97.7\% & \ -0.1\%\\
\hline
Ours: & &&&\\
GA($\lambda_{1}=0.3,\lambda_{2}=0.4,\lambda_{3}=0.3$) &\ 0.93\%\ &\textbf{\ 6.22\%}&\textbf{\ 94.30\%}&\ -0.13\%\\
GA($\lambda_{1}=0.5,\lambda_{2}=0.5,\lambda_{3}=0$) &\ 0.87\%\ & \textbf{\ 6.10}\%&\ 71.63\%&\ -0.07\%\\
GA($\lambda_{1}=0.5,\lambda_{2}=0,\lambda_{3}=0.5$) &\ 0.89\%\ & \ 9.00\%&\ \textbf{95.42\%}&\ -0.09\%\\
GA($\lambda_{1}=0.8,\lambda_{2}=0.1,\lambda_{3}=0.1$) &\textbf{\ 0.85\%\ }&\ 8.16\%&\ 91.00\%&\ -0.05\%\\
\hline
\end{tabular}
\end{table}
Compared with other approaches, albeit we do not obtain a minimal sparsity, our computation achieves outstanding reduction because of coarse granularity pruning. While some approaches with larger sparsity always employ fine granularity pruning, which is very tractable for facilitating sparsity but not essentially reducing the FLOPs of sparse weight tensors. Furthermore, our approach can perform a multi-objective trade-off according to the actual requirements whereas state-of-the-art approaches are unable to achieve this task.
\section{Conclusion}
We propose the heuristic GA to prune CNNs based on the multi-objective trade-off, which can obtain a variety of desirable compression performances. Moreover, we develop a two-step pruning framework for evolutionary algorithms, which may open a door to introduce the biological-inspired methodology to the field of CNNs pruning. As a future work, GA will be further investigated and improved to prune more large-scale CNNs.
%
%

%
%
%
%

\end{document}